\pdfoutput=1

\documentclass[11pt]{article}

\usepackage[final]{acl}

\usepackage{times}
\usepackage{latexsym}
\usepackage{amsmath}
\usepackage{amssymb}
\usepackage{algorithm}
\usepackage{algorithmicx}
\usepackage{algpseudocode}
\usepackage{multirow}
\usepackage{adjustbox}
\usepackage{booktabs}
\usepackage{colortbl}
\usepackage{xcolor}
\usepackage{graphicx} 
\usepackage{tikz}
\usepackage{pgfplots}
\pgfplotsset{compat=1.7}
\usetikzlibrary{pgfplots.groupplots}
\usetikzlibrary{shapes.geometric}
\usepackage{pgfplotstable}
\pgfplotsset{compat=newest}
\usetikzlibrary{pgfplots.fillbetween}
\usepackage{makecell}
\usepackage[T1]{fontenc}

\usepackage[utf8]{inputenc}
\usepackage{cleveref}

\usepackage{microtype}

\usepackage{inconsolata}

\usepackage{graphicx}
\usepackage{xcolor}
\title{LPOI: Listwise Preference Optimization for Vision Language Models} 

\author{
 \textbf{Fatemeh Pesaran zadeh\textsuperscript{1}} \quad
 \textbf{Yoojin Oh\textsuperscript{1}} \quad 
 \textbf{Gunhee Kim\textsuperscript{1}$^*$}\quad
 \\
 \textsuperscript{1}Seoul National University,
\\
\texttt{\small fatemehpesaran@vision.snu.ac.kr, snujin20@snu.ac.kr } \\
\texttt{\small gunhee@snu.ac.kr} 
}
\newcommand{\correspondingfootnote}{
    \let\oldthefootnote=\thefootnote
    \renewcommand{\thefootnote}{}
    \footnotetext{$*$ Corresponding author.}
    \let\thefootnote=\oldthefootnote
}

\begin{document}
\maketitle
\begin{abstract}
Aligning large VLMs with human preferences is a challenging task, as methods like RLHF and DPO often overfit to textual information or exacerbate hallucinations.
Although augmenting negative image samples partially addresses these pitfalls, no prior work has employed listwise preference optimization for VLMs, due to the complexity and cost of constructing listwise image samples.
In this work, we propose LPOI, the first object-aware listwise preference optimization developed for reducing hallucinations in VLMs.
LPOI identifies and masks a critical object in the image, and then interpolates the masked region between the positive and negative images to form a sequence of incrementally more complete images.
The model is trained to rank these images in ascending order of object visibility, effectively reducing hallucinations while retaining visual fidelity.
LPOI requires no extra annotations beyond standard pairwise preference data, as it automatically constructs the ranked lists through object masking and interpolation.
Comprehensive experiments on MMHalBench, AMBER, and Object HalBench confirm that LPOI outperforms existing preference optimization methods in reducing hallucinations and enhancing VLM performance. We make the code available at \url{https://github.com/fatemehpesaran310/lpoi}.
\end{abstract}

\correspondingfootnote

\section{Introduction}

\begin{figure}[t]
    \includegraphics[width=0.48\textwidth]{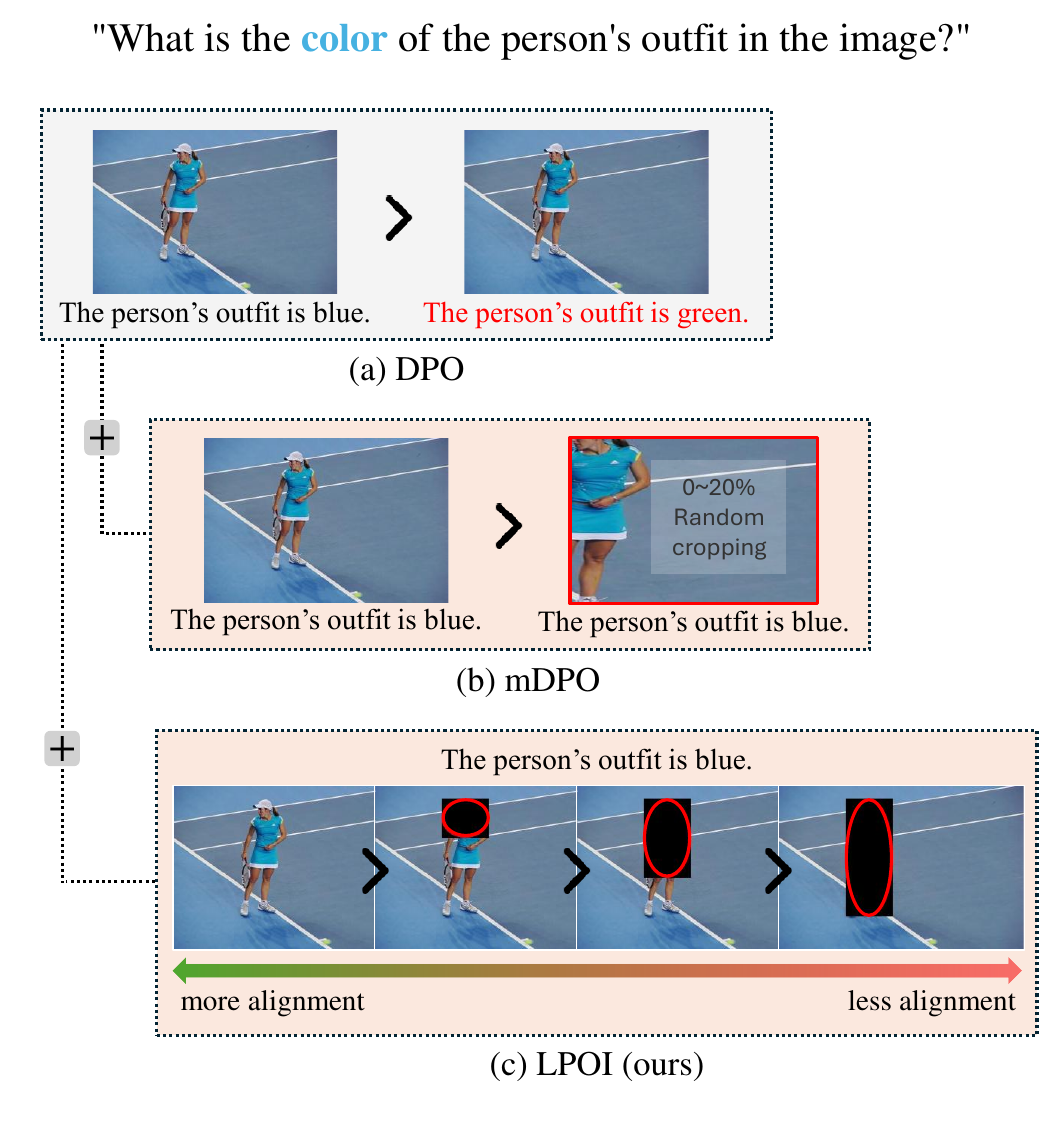}
    \caption{Comparison of preference optimization (PO) strategies for VLMs, with text and image negatives backgrounded in \textcolor{gray}{gray} and \textcolor{orange}{orange}, respectively. (a) DPO \cite{rafailov2024directpreferenceoptimizationlanguage}: PO with text negatives. (b) mDPO \cite{wang2024mdpoconditionalpreferenceoptimization}: DPO + PO using randomly cropped images as binary image negatives. (c) The proposed LPOI method: DPO + listwise PO with ranked image negatives, consisting of four samples: (1) the full image, (2) an image with the partial outfit, (3) an image with no outfit but some parts of person, and (4) an image with neither outfit nor person.}
    \label{fig:small}
\end{figure}

Aligning large language models (LLMs) or vision language models (VLMs) with human preferences has been an emergent challenge in the field. Approaches like Reinforcement Learning with Human Feedback (RLHF) \cite{ouyang2022traininglanguagemodelsfollow, glaese2022improvingalignmentdialogueagents, bai2022traininghelpfulharmlessassistant, stiennon2022learningsummarizehumanfeedback} and Direct Preference Optimization (DPO) \cite{rafailov2024directpreferenceoptimizationlanguage, li2023silkiepreferencedistillationlarge} have increasingly tackled this problem in the text domain. However, adapting these methods to multimodal settings introduces substantial challenges; simply substituting textual preference data with multimodal ones often leads to unreliable results and can even amplify critical issues like hallucinations \cite{zhao2024hallucinationsenhancinglvlmshallucinationaware, yue2024moremitigatingmultimodalhallucination}. 

In this regard, a line of research has revealed that multimodal models often overfit to textual information in the preference data, overlooking the necessary information in the image \cite{wang2024mdpoconditionalpreferenceoptimization,xie2024vdpomitigatinghallucinationlarge}. They propose augmenting the negative samples in the preference data via randomly cropping the image or editing the image using diffusion models.

Meanwhile, recent studies have demonstrated that the methods employing listwise samples for preference optimization often surpass the ones  based on pairwise samples by directly optimizing the entire ranking order in the list \cite{inproceedings, wu2019sqlranklistwiseapproachcollaborative, li2023integratinglistwiserankingpairwisebased}. This approach can capture interdependencies among items, unlike pairwise ranking that only compares two items at a time. 
Although efforts have been made to adapt DPO to listwise ranking in the text domain \cite{bansal2024comparingbadapplesgood, liu2024lipolistwisepreferenceoptimization, song2024preferencerankingoptimizationhuman, yuan2023rrhfrankresponsesalign}, applying this to images remains unexplored due to the complexity of ranking visual data and high cost of collecting listwise image samples.

\begin{figure*}[t]
    \centering
    \includegraphics[width=1.0\textwidth]{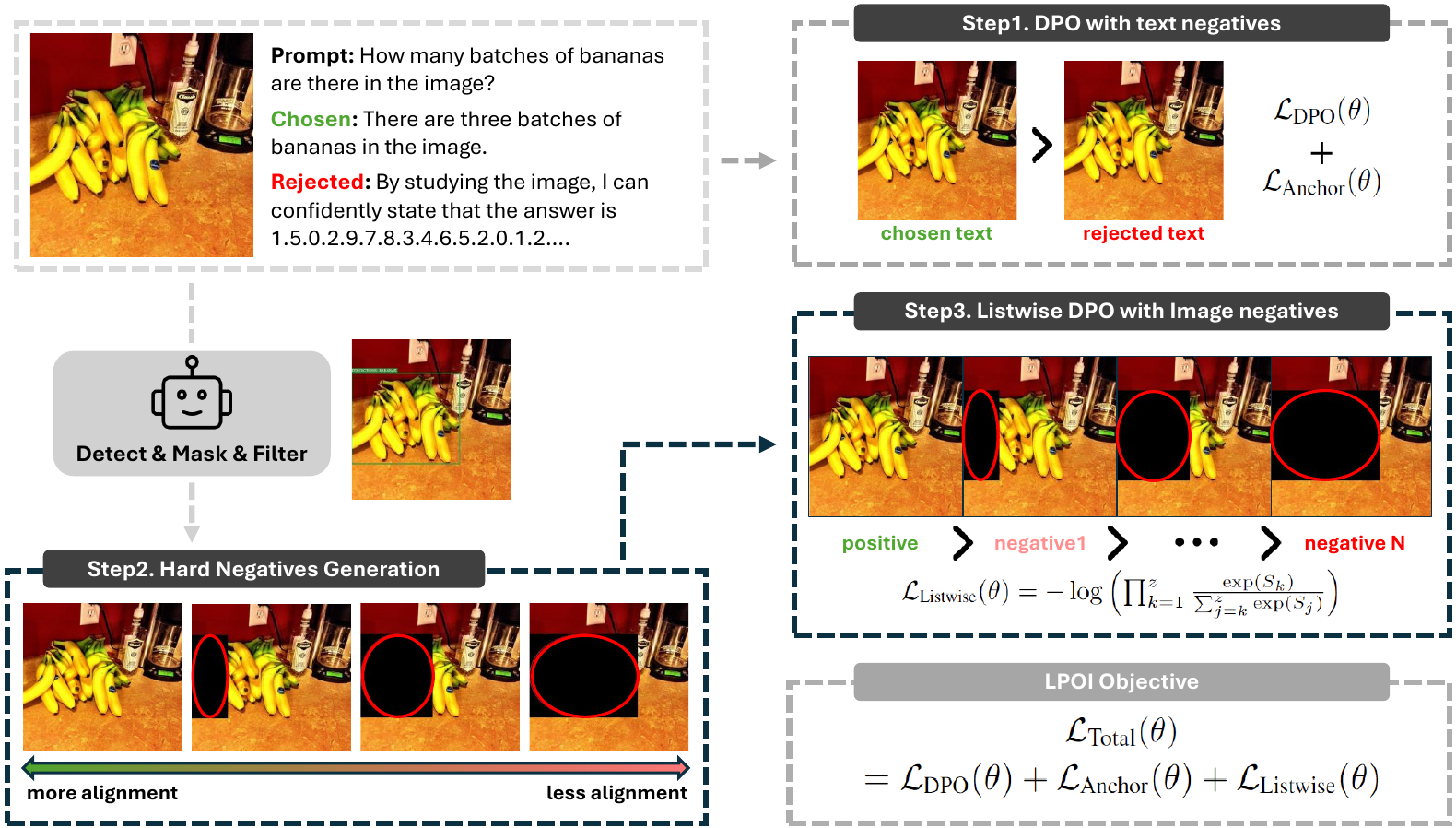}
    \caption{Overview of the LPOI framework. (1) Given an input image, prompt and corresponding set of chosen and rejected responses, we first compute $L_{DPO}$ and $L_{Anchor}$ using the response pairs similar to traditional DPO. (2) An object detection model and a VLM are employed to identify the most important object in the image. These objects are progressively masked in a sequence, with more visual clues being masked as the image deviates further from the positive example. (3) We optimize our model using this sequence of progressively masked images, which allows it to better differentiate between varying levels of hallucination, thereby improving its ability to discern subtle changes in visual context and generate responses more accurately grounded in the image.}
    \label{fig:pipeline}
\end{figure*}

To address this, we propose LPOI (Listwise Preference Optimization via Interpolating between Images), an object-aware listwise preference optimization framework for reducing hallucinations in VLMs. LPOI begins by identifying the critical object in an image based on textual context and creating \textit{hard negative} images by masking this object while keeping overall context. Next, LPOI interpolates the masking ratios between the positive and hard negative images, automatically generating a preference list to be optimized (\Cref{fig:pipeline}). Finally, the model is trained to rank these interpolated images using a listwise preference loss.

LPOI ranks images by how much of a critical object mentioned in the associated text they reveal (\Cref{fig:small}). Thus, the model's likelihood of generating positive text about the object increases with its visibility. By aligning the model’s output with the object’s actual presence, LPOI can lower hallucination rates compared to the state-of-the-art VLM preference optimization approaches. We also employ visual prompting~\cite{shtedritski2023doesclipknowred} to highlight the masked region in each negative example, redirecting the model’s focus to the missing object (\Cref{fig:saliency}). By efficiently generating diverse image lists without costly annotations or diffusion models, LPOI helps the model learn subtle distinctions between factual and hallucinating text, learning more robust and nuanced representation.

To empirically evaluate LPOI's reduction of hallucination, we fine-tune three VLM models, Idefics-8B \citep{laurençon2024mattersbuildingvisionlanguagemodels}, LLaVA-v1.5-7B, and LLaVA-v1.5-13B \citep{liu2024improvedbaselinesvisualinstruction}, and assess their performance on the MMHalBench \cite{sun2023aligninglargemultimodalmodels}, AMBER \cite{wang2024amberllmfreemultidimensionalbenchmark}, and Object HalBench \cite{rohrbach2019objecthallucinationimagecaptioning}. Our experiments demonstrate that preference learning in multimodal model benefits from the use of incrementally ranked listwise negatives, in reducing hallucinations and improving overall model performance.  

Our contributions can be outlined as follows.
\begin{itemize} \itemsep=0pt
    \item We present LPOI, the first approach to apply listwise ranking for VLM preference optimization to reduce hallucinations without requiring additional annotation beyond standard pairwise preference data. This is achieved by masking the image's critical object, and then interpolating the mask ratios between positive and negative images to generate the preference list automatically.
    \item We evaluate LPOI with three VLM models across three hallucination benchmarks. The results show that LPOI consistently achieves a lower hallucination rate compared to state-of-the-art VLM preference learning methods. Furthermore, LPOI outperforms existing methods in various scenarios, including when trained on datasets of different sizes or compared under a fixed budget of GPU hours.
\end{itemize}

\section{Related Work}

\textbf{Preference Learning.}
Aligning LLMs or VLMs with human preferences and values, known as preference learning, is an emerging challenge. Reinforcement learning with human feedback (RLHF) typically involves a multi-phase pipeline, including supervised fine-tuning of the policy model, training a reward model, and optimizing the policy based on the reward model \cite{christiano2023deepreinforcementlearninghuman,ouyang2022traininglanguagemodelsfollow, ziegler2020finetuninglanguagemodelshuman, gao2022scalinglawsrewardmodel, zadeh2024text2chart31instructiontuningchart}. Direct Preference Optimization (DPO) \cite{rafailov2024directpreferenceoptimizationlanguage} has emerged as a promising alternative, demonstrating remarkable performance while simplifying the process by eliminating the need for reward model training.
Following the DPO, numerous works have been proposed to enhance preference alignment for LLMs \cite{hong-etal-2024-orpo,xu2024thingscringeothersiterative,meng2024simposimplepreferenceoptimization,xu2024contrastivepreferenceoptimizationpushing}.


\paragraph{Preference Learning for VLMs.}
  Several studies have focused on adapting DPO to VLMs, primarily by constructing preference datasets \cite{xiao2024detectingmitigatinghallucinationlarge,zhou2024aligningmodalitiesvisionlarge, pi2024strengtheningmultimodallargelanguage,deng2024enhancinglargevisionlanguage}. Other approaches have explored generating negative images and using them in preference learning, either through random cropping \cite{wang2024mdpoconditionalpreferenceoptimization} or using computationally expensive diffusion models \cite{xie2024vdpomitigatinghallucinationlarge}. 
  In this work, we propose automatically generating hard negative samples by identifying the critical objects in the image using an object detection module and textual information, and then masking these objects out of the original image.


\paragraph{Hard Negative Mining.}
Hard negative mining has been extensively explored in deep metric learning and contrastive learning, with techniques like contrastive loss \cite{1640964}, triplet loss \cite{Schroff_2015}, and adaptive sampling \cite{robinson2021contrastivelearninghardnegative}.
They aim to enhance representation learning by identifying challenging negatives that are semantically close to positive samples. In our work, we adapt this principle to create hard negative images by preserving the overall semantic context of the image while masking out the critical object.

\paragraph{Listwise Ranking.}
Empirical and mathematical studies have shown that listwise ranking is more effective than pairwise ranking \cite{inproceedings,li2023integratinglistwiserankingpairwisebased,wu2019sqlranklistwiseapproachcollaborative}, since it optimizes the entire ranked list simultaneously, considering the relative positions of all items within the list. 
While prior work has focused on adapting DPO for listwise ranking in text-based applications \cite{bansal2024comparingbadapplesgood, liu2024lipolistwisepreferenceoptimization, song2024preferencerankingoptimizationhuman, yuan2023rrhfrankresponsesalign}, adapting listwise ranking in the VLM domain remains underexplored due to the high costs associated with collecting listwise image preference data. 
Our approach is the first to effectively leverage listwise ranking for VLM preference optimization to reduce hallucinations, without incurring additional annotation costs.

\begin{algorithm*}
\caption{Listwise Preference Optimization via Interpolating between Images (LPOI)}
\begin{algorithmic}[1]
\small
\Require Policy network $\pi_{\theta}$, reference policy network $\pi_{\text{ref}}$, dataset $\mathcal{D}$, parameters $N$, list size $L$
\For{$i = 1$ to $N$}
    \State Sample $(x, q, w, l) \sim \mathcal{D}$ \Comment{$x$: input image, $q$: question, $w$: chosen answer, $l$: rejected answer}
    \State Calculate $
        \mathcal{L}_{\text{DPO}}(\theta) = -\log \sigma \left( \beta \log \frac{\pi_{\theta}(w\,|\,x, q)}{\pi_{\text{ref}}(w\,|\,x, q)} - \beta \log \frac{\pi_{\theta}(l\,|\,x, q)}{\pi_{\text{ref}}(l\,|\,x, q)} \right)
    $
    \State Calculate $
        \mathcal{L}_{\text{Anchor}}(\theta) = -\log \sigma \left( \beta \log \frac{\pi_{\theta}(w\,|\,x, q)}{\pi_{\text{ref}}(w\,|\,x, q)} - \delta \right)
    $

    \State Extract bounding box of the main object $b$ from $x$, prompt $q$ and chosen answer $w$. \label{line:5}
    \For{$k = 1$ to $L$} \Comment{Create $k$-th negative sample in the list}
        \State Define $m_k$ as the mask obtained by masking $\left(\frac{k-1}{L-1}\right) \times 100\%$ of the bounding box $b$.
        \State $x_k = \text{Highlight}(\text{Mask}(x, m_k))$ \Comment{Apply masking and visual prompting}
    \EndFor
    \If{filtering model answers $(x_L, q, w)$ to be positive answer}
    \State Go to Line 5 with different object $b$
    \EndIf \label{line:12}
    \State Calculate $
            \mathcal{L}_{\text{Listwise}}(\theta) = -\log \left( \prod_{k=1}^z \frac{\exp(S_k)}{\sum_{j = k}^z \exp(S_j)} \right)
        $
    where $
        S_k = \beta \log \frac{\pi_{\theta}(w\,|\,x_k, q)}{\pi_{\text{ref}}(w\,|\,x_k, q)}
    $
    \State Minimize $\mathcal{L}_{\text{Total}}(\theta) = \mathcal{L}_{\text{DPO}}(\theta) + \mathcal{L}_{\text{Anchor}}(\theta) + \mathcal{L}_{\text{Listwise}}(\theta)$ \Comment{Optimize towards $S_1 > S_2 > \cdots > S_L$}
\EndFor
\end{algorithmic}
\label{alg:LPOI}
\end{algorithm*}

\section{Approach}


A major challenge in preference learning for VLMs is that models often overfit to textual patterns and overlook the image information \citep{wang2024mdpoconditionalpreferenceoptimization}. This issue can lead to object hallucination \cite{rohrbach2019objecthallucinationimagecaptioning}, where the model erroneously describes objects or attributes that do not actually appear in the visual scene; particularly when there are no proper negative image samples during training.
In this work, we propose to reduce object hallucination by addressing two key objectives: (1) Generating hard negative image samples, in which the critical object mentioned in the text is missing but the overall context is preserved 
(\Cref{hardnegative generation}). (2) Creating listwise samples without any additional costly annotations, where the images are aligned with the object's actual presence. 
(\Cref{listwiseoptimization}).

\subsection{Hard Negative Sample Generation} \label{hardnegative generation}

We generate hard negative image samples—images that turn the originally preferred answer into the hallucinated one while preserving the overall semantic context—through two steps of detecting the object to be masked, and applying the mask (Figure \ref{fig:pipeline}). 
First, we run the zero-shot object detection module, Grounding-DINO-Tiny with 172M parameters \citep{liu2024groundingdinomarryingdino}, through the input image. We select the object to be masked in the following orders: objects in the first sentence of the chosen answer, then those in the query, and finally any remaining objects in the answer. We also randomly select a detected object that are not in the text. For the selected object, we mask its bounding box and highlight it using a visual prompting technique (e.g., a red circle) \cite{shtedritski2023doesclipknowred}, directing the model’s attention to the masked area.

We then verify that the masked image is indeed a hard negative sample by making sure that Idefics2-8B \cite{laurençon2024mattersbuildingvisionlanguagemodels} hallucinates. If it does not hallucinate, another object is selected, and the process is repeated (Algorithm \ref{alg:LPOI}, Lines \ref{line:5}–\ref{line:12}). 

\begin{table*}[t]
\centering
\small
\setlength{\tabcolsep}{2.0pt}
\begin{adjustbox}{max width=2.0\columnwidth}
\begin{tabular}{lcccccccc}\toprule
&\multicolumn{2}{c}{Object HalBench} &\multicolumn{2}{c}{MMHalBench}  &\multicolumn{4}{c}{AMBER}
 \\ \cmidrule(lr){2-3} \cmidrule(lr){4-5} \cmidrule(lr){6-9}
Method  &CHAIR$_{s}$ $\downarrow$&CHAIR$_{i}$ $\downarrow$ &Score $\uparrow$ &HalRate $\downarrow$ & CHAIR$_{s}$ $\downarrow$&Cover. $\uparrow$ &HalRate $\downarrow$&Cog. $\downarrow$ \\
 \midrule
LLaVA-v1.5-7B \cite{liu2024improvedbaselinesvisualinstruction} & 49.7 & 26.1 & 2.02 & 0.65 & 7.7 & 49.8 & 31.9 & 3.7 \\
+ DPO \citep{rafailov2024directpreferenceoptimizationlanguage}  & 42.3 & 23.2 & 2.00 & 0.69& 6.7 & \textbf{53.2} & 33.7& 3.3 \\
+ HALVA \cite{sarkar2024dataaugmentedphraselevelalignmentmitigating} &  -& -& - & - & 6.6 & 53.0 & 32.2 & 3.4 \\
+ HA-DPO \cite{zhao2024hallucinationsenhancinglvlmshallucinationaware}  & 39.9 & 19.9 & - & - & 6.7 & 49.8 & 30.9 & 3.3 \\
+ V-DPO \cite{xie2024vdpomitigatinghallucinationlarge} & - & - & - & - & 6.6 & 49.1 & 30.8 & 3.1 \\
+ mDPO \cite{wang2024mdpoconditionalpreferenceoptimization} &  30.7 & 16.0 & \textbf{2.40} & \textbf{0.59} & 5.0 & 52.5 & 27.5 & 2.4 \\
+ LPOI (Ours) &  \textbf{24.3} & \textbf{14.6} & \textbf{2.40} & \textbf{0.59} & \textbf{4.3} & 51.9 & \textbf{26.4} & \textbf{2.0} \\
\midrule
LLaVA-v1.5-13B \cite{liu2024improvedbaselinesvisualinstruction}& 44.3 & 21.2 & 2.09 & 0.64 & 6.3 & 51.0 & 30.2 & 3.0 \\
+ DPO \citep{rafailov2024directpreferenceoptimizationlanguage} & 38.3 & 19.4 & 2.36 & 0.61 & 6.2 & \textbf{54.3} & 31.8 & 2.6 \\
 + mDPO \citep{wang2024mdpoconditionalpreferenceoptimization} &  33.3 & 16.6 & 2.50 & \textbf{0.57} &4.6 & 52.6 & 25.0 & 2.0 \\
 + LPOI (Ours) & \textbf{24.3} & \textbf{11.7} & \textbf{2.54} & \textbf{0.57} & \textbf{3.9} & 52.9 & \textbf{22.3} & \textbf{1.8}\\
\midrule
Idefics2-8B \cite{laurençon2024mattersbuildingvisionlanguagemodels}  & 6.3 & 4.2 & 2.62 & 0.43& 3.4 & 36.5 & 7.6 & 0.4 \\
+ DPO \citep{rafailov2024directpreferenceoptimizationlanguage} & 6.0 & 4.2 & 2.48 & 0.45 & 3.5 & 37.4& 8.1 & \textbf{0.2} \\
+  mDPO \citep{wang2024mdpoconditionalpreferenceoptimization} &7.3 & 5.4 & 2.80 &0.40 & 2.7 & \textbf{37.7} & 6.2 &\textbf{0.2} \\
+ LPOI (Ours) & \textbf{5.3} & \textbf{3.6} & \textbf{2.88} & \textbf{0.36} & \textbf{2.6} & 36.4 & \textbf{5.7} & \textbf{0.2}\\
\bottomrule
\end{tabular}
\end{adjustbox}
\caption{Performance comparison between various preference learning methods on  Object HalBench, MMHalBench, and AMBER benchmarks. We use three base VLM models: Llava-v1.5-7B/13B and Idefics2-8B. The results of DPO and mDPO are reproduced under a fair setting with LPOI. HALVA, HA-DPO, and V-DPO are taken from their respective papers; they are included for reference.}
\label{tab:main}
\end{table*}

\subsection{Listwise Optimization} \label{listwiseoptimization}

We automatically create listwise samples with no annotation 
by interpolating the masking ratios between the positive image and the hard negative image. Specifically, when generating $k$-th image in the list, we progressively mask $\frac{k-1}{L-1} \times 100\%$ of the bounding box starting from the side closest to the image edge, where $L$ denotes the list size. As a result, we obtain a list of samples aligned by the visibility, where images with less masking are more positive and those with more masking are more negative.


Once the listwise samples are created, we optimize the model to have higher likelihood of generating positive response according to the order of the list. This is achieved by using a listwise ranking loss, which can be interpreted as the negative log-likelihood of a given permutation \citep{Cao2007LearningTR,rafailov2024directpreferenceoptimizationlanguage,liu2024lipolistwisepreferenceoptimization}: 

\begin{align}
\label{eq:listwise}
\mathcal{L}_{\text{Listwise}}(\theta) = -\log \left( \prod_{k=1}^z \frac{\exp(S_k)}{\sum_{j = k}^z \exp(S_j)} \right),
\end{align}
where $S_k = \beta \log \frac{\pi_{\theta}(w\,|\,x_k, q)}{\pi_{\text{ref}}(w\,|\,x_k, q)}$. 
Here, $\pi_\theta$ and $\pi_{\text{ref}}$ denote the fine-tuned model and the base model, respectively. $S_k$ is the normalized log-likelihood of the model $\pi_\theta$ describing the relevant object given the image $x_k$. $x_1$ is the original image, $x_L$ is the hard negative image, and $x_k$ is the interpolated image with the masking ratio of $\frac{k-1}{L-1} \times 100\%$.

By minimizing the listwise loss in \cref{eq:listwise}, we optimize the values of $S_k$ to be $S_1 > S_2 > \cdots > S_L$, which implies that the model's likelihood of generating positive text about the object increases as its visibility in the image grows (\Cref{fig:pipeline}). This approach helps the model reduce hallucinations, as it encourages the model to mention the object in proportion to its visibility.

In addition to the listwise loss, we also use the standard DPO loss and the anchor loss:
$$\mathcal{L}_{\text{Anchor}}= -\log \sigma \left( \beta \log \frac{\pi_{\theta}(w\,|\,x, q)}{\pi_{\text{ref}}(w\,|\,x, q)} - \delta \right),$$ 
which is proposed in mDPO \cite{wang2024mdpoconditionalpreferenceoptimization}. Minimizing the anchor loss further increases the likelihood that the model generates postive responses when given the original image. 
In total, our objective becomes 
\begin{align*}
\mathcal{L}_{\text{Total}}(\theta) = \mathcal{L}_{\text{DPO}}(\theta) + \mathcal{L}_{\text{Anchor}}(\theta) + \mathcal{L}_{\text{Listwise}}(\theta).
\end{align*}
Algorithm \ref{alg:LPOI} summarizes the overall procedure of the proposed LPOI method.

\section{Experiment}

\begin{table*}[t]
\centering
\small
\setlength{\tabcolsep}{3pt} 
\begin{adjustbox}{max width=2.5\columnwidth}
\begin{tabular}{lccccccccc}\toprule
 & \multicolumn{2}{c}{Object HalBench} & \multicolumn{2}{c}{MMHalBench} & \multicolumn{4}{c}{AMBER} \\ 
\cmidrule(lr){2-3} \cmidrule(lr){4-5} \cmidrule(lr){6-9}
Method & CHAIR$_{s}$ $\downarrow$ & CHAIR$_{i}$ $\downarrow$ & Score $\uparrow$ & HalRate $\downarrow$ & CHAIR$_{s}$ $\downarrow$& Cover. $\uparrow$ & HalRate $\downarrow$ & Cog. $\downarrow$ \\ \midrule
Idefics2-8B \citep{laurençon2024mattersbuildingvisionlanguagemodels} & 6.3 & 4.2 & 2.62 & 0.43 & 3.4 & 36.5 & 7.6 & 0.4 \\
+ DPO \citep{rafailov2024directpreferenceoptimizationlanguage} & 6.0 & 4.3 & 2.29 & 0.51 & 3.1 & 36.4 & 6.8 & \textbf{0.3} \\ 
+ mDPO \citep{wang2024mdpoconditionalpreferenceoptimization} & 8.7 & 5.6 & 2.71 & 0.42 & \textbf{2.8} & \textbf{37.2} & 6.5 & \textbf{0.3} \\
+ LPOI (Ours) & \textbf{5.3} & \textbf{4.0} & \textbf{2.81} & \textbf{0.38} & \textbf{2.8} & 36.2 & \textbf{6.2} & \textbf{0.3}\\
\bottomrule
\end{tabular}
\end{adjustbox}
\caption{Performance comparison under the same training cost (20 hours on a single RTX A6000 GPU) for Idefics2-8B model on Object HalBench, MMHalBench, and AMBER benchmarks.}
\label{tab:table4}
\end{table*}

\begin{table}[t]
\centering
\small
\setlength{\tabcolsep}{3pt} 
\begin{adjustbox}{max width=1.0\columnwidth}
\begin{tabular}{lccccccc}\toprule
 & \multicolumn{2}{c}{Object HalBench} & \multicolumn{2}{c}{MMHalBench} & \multicolumn{2}{c}{AMBER} \\ 
\cmidrule(lr){2-3} \cmidrule(lr){4-5} \cmidrule(lr){6-7}
Method & CHAIR$_{s}$ $\downarrow$ & CHAIR$_{i}$ $\downarrow$ & Score $\uparrow$ & HalRate $\downarrow$ & CHAIR$_{s}$ $\downarrow$ & HalRate $\downarrow$ \\ \midrule
without V.P. & 5.3 & 4.0 & 2.74& 0.40 & 2.7 & 6.0\\
with V.P. & \textbf{5.0} & \textbf{3.4} & \textbf{2.91} & \textbf{0.35} & \textbf{2.6} & \textbf{5.8} \\
\bottomrule
\end{tabular}
\end{adjustbox}
\caption{Performance comparison with and without visual prompting for the Idefics2-8B model on Object HalBench, MMHalBench, and AMBER benchmarks.}
\label{tab:table2}
\end{table}

\begin{table}[t]
\centering
\small
\setlength{\tabcolsep}{3pt} 
\begin{adjustbox}{max width=1.0\columnwidth}
\begin{tabular}{lccccccc}\toprule
 & \multicolumn{2}{c}{Object HalBench} & \multicolumn{2}{c}{MMHalBench} & \multicolumn{2}{c}{AMBER} \\ 
\cmidrule(lr){2-3} \cmidrule(lr){4-5} \cmidrule(lr){6-7}
Method &  CHAIR$_{s}$ $\downarrow$ & CHAIR$_{i}$ $\downarrow$ & Score $\uparrow$ & HalRate $\downarrow$ & CHAIR$_{s}$ $\downarrow$ & HalRate $\downarrow$ \\ \midrule
List size 3  & 7.3 & 5.1 & 2.86 & \textbf{0.36} & 2.9 & 6.6 \\
List size 4  & 6.7 & 4.5 & 2.86 & \textbf{0.36} & \textbf{2.5} & \textbf{5.6} \\
List size 5  & \textbf{5.3} & \textbf{3.6} & \textbf{2.88} & \textbf{0.36} & 2.6 & 5.7 \\
\bottomrule
\end{tabular}
\end{adjustbox}
\caption{Performance comparison across different list sizes for the Idefics2-8B model on Object HalBench, MMHalBench, and AMBER benchmarks.}
\label{tab:table3}
\end{table}

\subsection{Experimental Setup}
\paragraph{Baselines.}
We compare our LPOI approach against established methods, including DPO \cite{rafailov2024directpreferenceoptimizationlanguage}, mDPO \cite{wang2024mdpoconditionalpreferenceoptimization}, HALVA \cite{sarkar2024dataaugmentedphraselevelalignmentmitigating}, HA-DPO \cite{zhao2024hallucinationsenhancinglvlmshallucinationaware}, and V-DPO \citep{xie2024vdpomitigatinghallucinationlarge}.
We evaluate each method using three VLMs including the LLaVA-v1.5-7B, LLaVA-v1.5-13B \cite{liu2024improvedbaselinesvisualinstruction}, and Idefics2-8B \cite{laurençon2024mattersbuildingvisionlanguagemodels}. For DPO and mDPO, we report reproduced results using the same training dataset as our LPOI method. For HALVA, HA-DPO, and V-DPO, we report the originally published performance for reference. 

\paragraph{Evaluation.}
We evaluate both the base and fine-tuned versions of VLMs using MMHalBench \cite{sun2023aligninglargemultimodalmodels}, Object HalBench \cite{rohrbach2019objecthallucinationimagecaptioning}, and AMBER \cite{wang2024amberllmfreemultidimensionalbenchmark}, which are standard benchmarks for assessing hallucination and the quality of generated text of VLMs. We report the CHAIR metric \cite{rohrbach2019objecthallucinationimagecaptioning} to measure object hallucination and the MMHalBench score (computed via GPT-4o \cite{openai2024gpt4ocard}) to quantify the quality of generated outputs.

\paragraph{Training setup.}
We conduct the preference learning via LoRA fine-tuning \cite{hu2021loralowrankadaptationlarge}. For training sets, we randomly sample 10K preference data from Silkie \cite{li2023silkiepreferencedistillationlarge} and instruction datafrom LLaVA-Instruct-150K \cite{liu2023visualinstructiontuning}, following the setup of mDPO \cite{wang2024mdpoconditionalpreferenceoptimization}. Idefics2-8B is trained for 3 epochs with a learning rate of 5e-6, and LLaVA-v1.5 (7B and 13B) for 1 epoch with a learning rate of 1e-6. We employ 1 RTX A6000 GPU for fine-tuning Idefics2-8B and LLaVA-v1.5-7B, and employ 2 RTX A6000 GPU for LLaVA-v1.5-13B. Refer to \Cref{sec:detail} for details on hyperparameters. 

\subsection{Results}
We present the results in \Cref{tab:main}. Our proposed LPOI consistently improves performance of different VLMs across most benchmarks. Notably, it excels at hallucination related metrics, including the HalRate in MMHalBench, the CHAIR metric in Object HalBench, and the CHAIR and cognition metric in AMBER. Specifically, our method achieves 24.3 in CHAIR$_s$  and 14.6 in CHAIR$_i$ for LLaVA-v1.5-7B on Object HalBench, which is superior than state-of-the-art mDPO with 30.7 in CHAIR$_s$ and 16.0 in CHAIR$_i$ in the same setting.
It is also worth noting that although our coverage performance is on par with other methods, this metric often grows at the expense of increased hallucination since it measures how much ratio of correct objects are detected by the model. Thus, models that generate more mentions, even if some are erroneous, can inflate their coverage score.

We further note that Object HalBench is generally more challenging than AMBER with respect to the CHAIR score, and models tend to exhibit a higher hallucination rate on this benchmark. Our method yields a notably larger performance gain on Object HalBench compared to AMBER, where models already maintain a low hallucination rate and the scores are largely saturated.

\subsection{Human Evaluation}
\begin{figure}[t]
    \centering
    \includegraphics[width=0.49\textwidth]{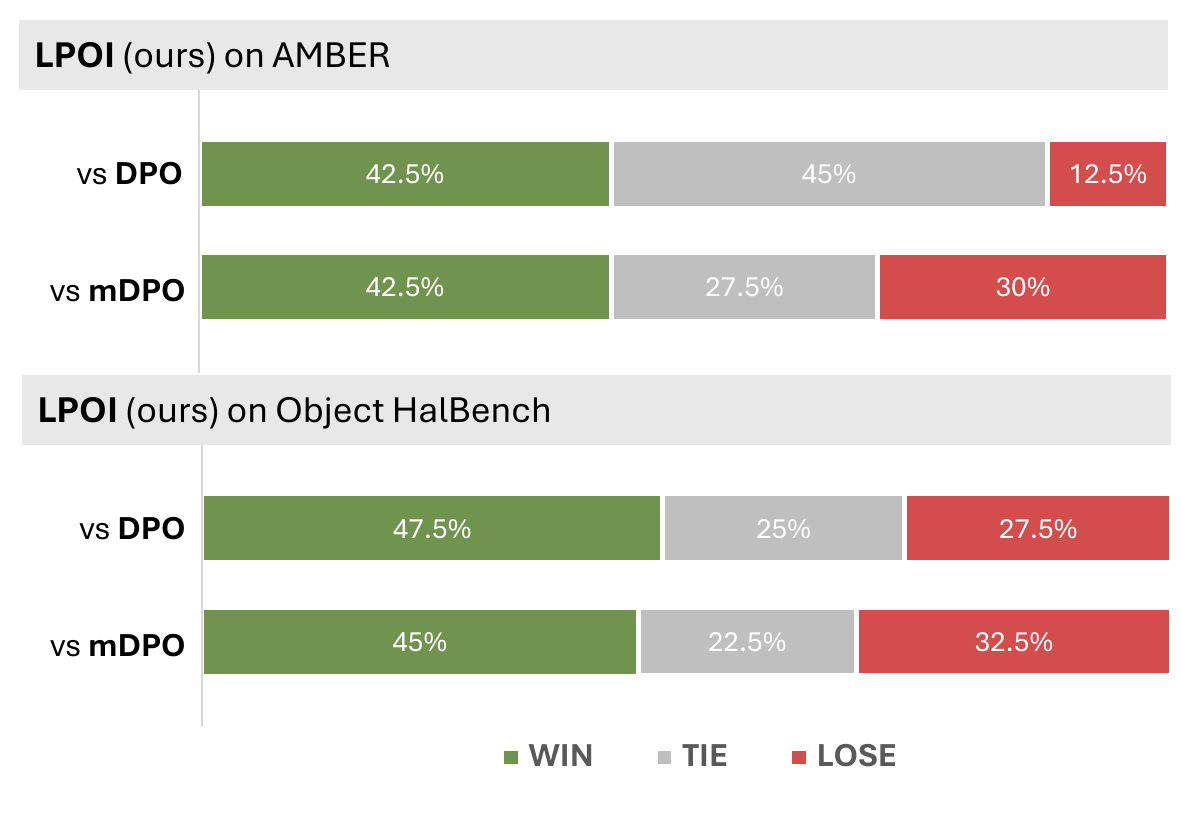}
    \caption{Human evaluation results on a subset of the AMBER and Object HalBench benchmark. We compare responses generated by the Idefics-2B model fine-tuned using LPOI (ours), DPO, and mDPO.}
    \label{fig:human}
\end{figure}

To further assess the quality of responses, we conduct a human evaluation using 80 randomly selected image-question pairs, 40 from the AMBER benchmark and 40 from the Object HalBench. We present the results in \Cref{fig:human}. Each pair is presented to three crowd workers recruited via Amazon Mechanical Turk from English-speaking countries, with a maximum payment of \$0.50 per HIT. The annotators are provided with two responses generated by the Idefics2-8B, one fine-tuned using our LPOI and the other using mDPO, which is the strongest baseline in \Cref{tab:main}. Workers are instructed to select the response that is more accurate and reliable, considering the visual information in the image.

We also compare with DPO under the same conditions. Annotators consistently prefer responses from our fine-tuned model over those from mDPO and DPO. Inter-annotator agreement is measured using Krippendorff’s $\alpha$, which yields a value of 0.735 for DPO and 0.671 for mDPO on the AMBER benchmark, and a value of 0.823 for DPO and 0.627 for mDPO on the Object HalBench. These values reflect the level of agreement among annotators regarding the relative quality of the responses, with three possible choices: A is better, B is better, or a tie. More details can be found in \Cref{sec:human}.

\begin{figure}[t]
    \centering
    \includegraphics[width=0.5\textwidth]{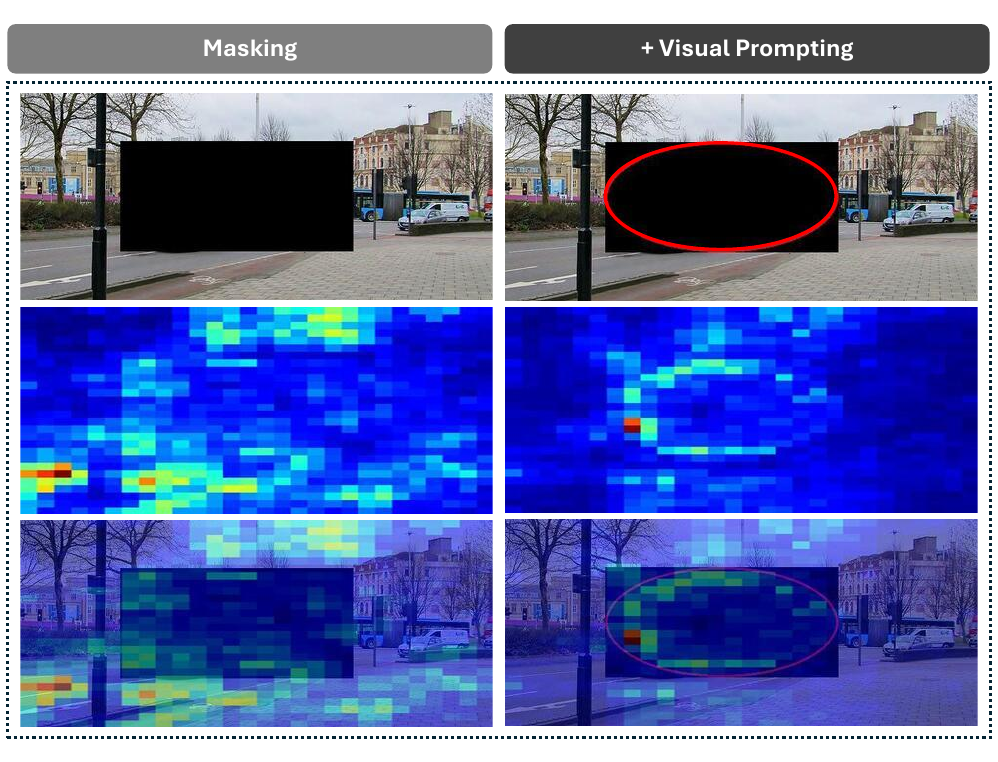}
    \caption{Comparison of saliency maps with or without visual prompting (highlighted in red circle). Visual prompting shifts the model’s attention towards the masked area, guiding it to focus more on the region of interest. In the saliency maps, blue indicates low saliency, while red indicates high saliency.}
    \label{fig:saliency}
\end{figure}

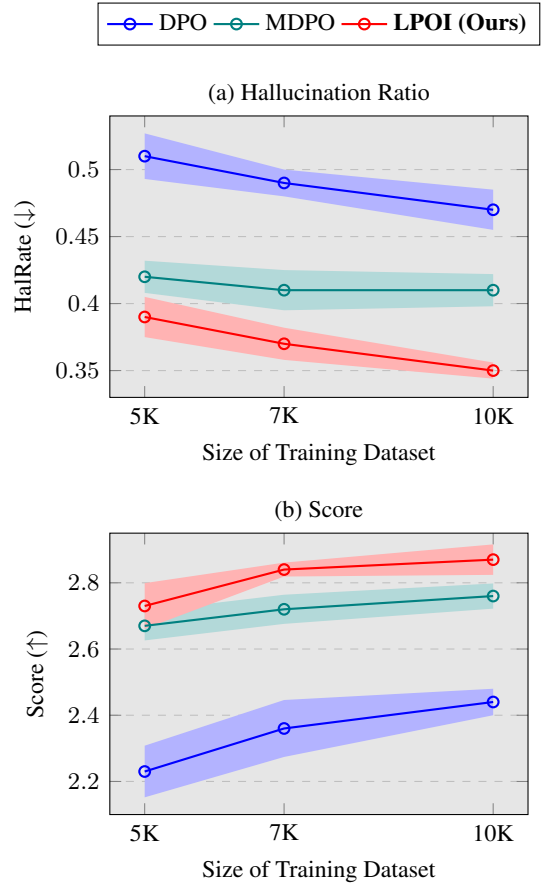
\begin{figure}[ht]
    \centering
    \begin{tikzpicture}
    \begin{groupplot}[
        group style={columns=1, rows=2, horizontal sep=0.0cm, vertical sep=1.8cm}, 
        width=0.92\columnwidth, 
        height=0.69\columnwidth,
        grid style={dash pattern=on 3pt off 2pt, line width=0.4pt}, 
        ymajorgrids=true,
        grid style=dashed,
        title style={at={(0.5,0.95)},anchor=south}, 
        legend style={at={(0.5,1.25)}, anchor=south, fill=none, draw=black, legend columns=3}, 
    ]

    \nextgroupplot[
        title={\small (a) Hallucination Ratio},
        xlabel={\small Size of Training Dataset},
        ylabel={\small HalRate ($\downarrow$)},
        xmin=4.5, xmax=10.5,
        ymin=0.33, ymax=0.54,
        xtick={5,7,10},
        xticklabels={5K, 7K, 10K},
        xticklabel style={font=\small},
        yticklabel style={font=\small},
        axis background/.style={fill=none},
        axis on top=false,
    ]

    \begin{pgfonlayer}{axis background}
      \fill[gray!20] (axis cs:4.500,0.33) rectangle (axis cs:10.500,0.45);
      \fill[gray!20] (axis cs:4.500,0.45) rectangle (axis cs:10.500,0.54);
    \end{pgfonlayer}

    \addplot[
        color=blue,
        mark=o,
        thick
    ] coordinates {
        (5,0.51)
        (7,0.49)
        (10,0.47)
    };
    \addlegendentry{\footnotesize{DPO}}

    \path[name path=dpoHigh]
        coordinate (A) at (5, {0.51+0.017})
        coordinate (B) at (7, {0.49+0.010})
        coordinate (C) at (10,{0.47+0.015});
    \draw[draw=none,name path=dpoHigh] (A) -- (B) -- (C);

    \path[name path=dpoLow]
        coordinate (D) at (5, {0.51-0.017})
        coordinate (E) at (7, {0.49-0.010})
        coordinate (F) at (10,{0.47-0.015});
    \draw[draw=none,name path=dpoLow] (D) -- (E) -- (F);

    \addplot[blue!30, forget plot] fill between[of=dpoHigh and dpoLow];

    \addplot[
        color=teal,
        mark=o,
        thick
    ] coordinates {
        (5,0.42)
        (7,0.41)
        (10,0.41)
    };
    \addlegendentry{\footnotesize{MDPO}}

    \path[name path=mdpoHigh]
        coordinate (G) at (5, {0.42+0.012})
        coordinate (H) at (7, {0.41+0.015})
        coordinate (I) at (10,{0.41+0.012});
    \draw[draw=none,name path=mdpoHigh] (G) -- (H) -- (I);

    \path[name path=mdpoLow]
        coordinate (J) at (5, {0.42-0.012})
        coordinate (K) at (7, {0.41-0.015})
        coordinate (L) at (10,{0.41-0.012});
    \draw[draw=none,name path=mdpoLow] (J) -- (K) -- (L);

    \addplot[teal!30, forget plot] fill between[of=mdpoHigh and mdpoLow];

    \addplot[
        color=red,
        mark=o,
        thick
    ] coordinates {
        (5,0.39)
        (7,0.37)
        (10,0.35)
    };
    \addlegendentry{\footnotesize{\textbf{LPOI (Ours)}}}

    \path[name path=hnlHigh]
        coordinate (M) at (5, {0.39+0.015})
        coordinate (N) at (7, {0.37+0.012})
        coordinate (O) at (10,{0.35+0.006});
    \draw[draw=none,name path=hnlHigh] (M) -- (N) -- (O);

    \path[name path=hnlLow]
        coordinate (P) at (5, {0.39-0.015})
        coordinate (Q) at (7, {0.37-0.012})
        coordinate (R) at (10,{0.35-0.006});
    \draw[draw=none,name path=hnlLow] (P) -- (Q) -- (R);

    \addplot[red!30, forget plot] fill between[of=hnlHigh and hnlLow];

    \nextgroupplot[
        title={\small (b) Score},
        xlabel={\small Size of Training Dataset},
        ylabel={\small Score ($\uparrow$)},
        xmin=4.5, xmax=10.5,
        ymin=2.10, ymax=2.95,
        xtick={5,7,10},
        xticklabels={5K, 7K, 10K},
        xticklabel style={font=\small},
        yticklabel style={font=\small},
        axis background/.style={fill=none},
        axis on top=false,
    ]

    \begin{pgfonlayer}{axis background}
      \fill[gray!20] (axis cs:4.500,2.10) rectangle (axis cs:10.500,2.95);
    \end{pgfonlayer}

    \addplot[
        color=blue,
        mark=o,
        thick
    ] coordinates {
        (5,2.23)
        (7,2.36)
        (10,2.44)
    };

    \path[name path=dpoHigh]
        coordinate (A) at (5, {2.23+0.078})
        coordinate (B) at (7, {2.36+0.086})
        coordinate (C) at (10,{2.44+0.040});
    \draw[draw=none,name path=dpoHigh] (A) -- (B) -- (C);

    \path[name path=dpoLow]
        coordinate (D) at (5, {2.23-0.078})
        coordinate (E) at (7, {2.36-0.086})
        coordinate (F) at (10,{2.44-0.040});
    \draw[draw=none,name path=dpoLow] (D) -- (E) -- (F);

    \addplot[blue!30, forget plot] fill between[of=dpoHigh and dpoLow];

    \addplot[
        color=teal,
        mark=o,
        thick
    ] coordinates {
        (5,2.67)
        (7,2.72)
        (10,2.76)
    };

    \path[name path=mdpoHigh]
        coordinate (G) at (5, {2.67+0.044})
        coordinate (H) at (7, {2.72+0.044})
        coordinate (I) at (10,{2.76+0.038});
    \draw[draw=none,name path=mdpoHigh] (G) -- (H) -- (I);

    \path[name path=mdpoLow]
        coordinate (J) at (5, {2.67-0.044})
        coordinate (K) at (7, {2.72-0.044})
        coordinate (L) at (10,{2.76-0.038});
    \draw[draw=none,name path=mdpoLow] (J) -- (K) -- (L);

    \addplot[teal!30, forget plot] fill between[of=mdpoHigh and mdpoLow];

    \addplot[
        color=red,
        mark=o,
        thick
    ] coordinates {
        (5,2.73)
        (7,2.84)
        (10,2.87)
    };

    \path[name path=hnlHigh]
        coordinate (M) at (5, {2.73+0.069})
        coordinate (N) at (7, {2.84+0.021})
        coordinate (O) at (10,{2.87+0.046});
    \draw[draw=none,name path=hnlHigh] (M) -- (N) -- (O);

    \path[name path=hnlLow]
        coordinate (P) at (5, {2.73-0.069})
        coordinate (Q) at (7, {2.84-0.021})
        coordinate (R) at (10,{2.87-0.046});
    \draw[draw=none,name path=hnlLow] (P) -- (Q) -- (R);

    \addplot[red!30, forget plot] fill between[of=hnlHigh and hnlLow];

    \end{groupplot}

    \end{tikzpicture}
    
    \caption{MMHalBench results for different preference optimization methods trained on three different sizes of training sets.}
    \label{fig:small-data}
\end{figure}

\definecolor{darkgreen}{rgb}{0.0, 0.8, 0.2}
\begin{figure*}[t]
    \centering
    \includegraphics[width=1.0\textwidth]{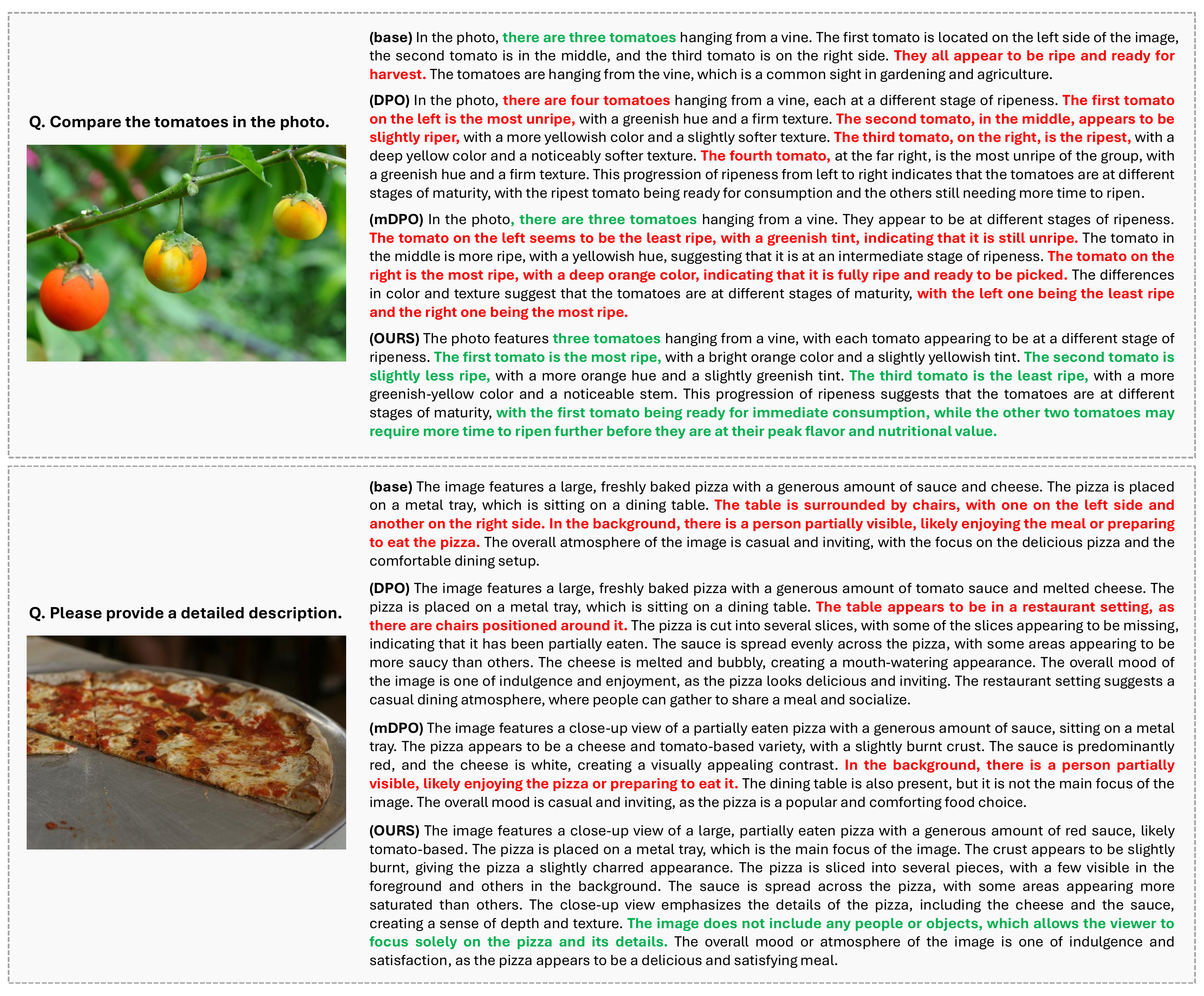}
    \caption{Qualitative results of the base model (LLaVA-v1.5-7B) its finetuned versions with DPO, mDPO, and LPOI (Ours). \textbf{\textcolor{darkgreen}{Correct answers}} and \textbf{\textcolor{red}{hallucinations}} are highlighted.}
    \label{fig:qualitativeexample}
\end{figure*}

\subsection{Analysis}

\paragraph{Comparison Under Equal Training Budget.}
We present the results of evaluating DPO, mDPO and LPOI (ours) under the same training budget (GPU hours). Since the listwise objective inherently incurs a higher training cost compared to the pairwise objective, we further present the results of training LPOI, DPO, and mDPO for 20 hours on a single RTX A6000 GPU using a 5K subsample of the preference dataset. Table \ref{tab:table4} demonstrates that, even under the same training budget, our method consistently outperforms DPO and mDPO, particularly in terms of hallucination scores and the overall quality of the generated outputs.

\paragraph{Advantages of Visual Prompting.}

Masking the critical object in an image may not always turn the original preferred answer into a negative one, when VLMs can still infer the correct answer by using surrounding context. Thus, we apply visual prompting \cite{shtedritski2023doesclipknowred, wu2024visualpromptingmultimodallarge, lin2024drawandunderstandleveragingvisualprompts, cai2024vipllavamakinglargemultimodal} to highlight more the masked region and guide the model's attention there. We validate that visual prompting directs the model's focus and increase the performance.

Figure \ref{fig:saliency} shows the saliency maps of the masked image with and without visual prompting. The saliency maps are computed using a gradient-based method with ResNet-50 \citep{he2015deepresiduallearningimage}, where we aggregate the absolute gradient of the prediction loss with respect to the input image across RGB channels. After the masked area is highlighted with a red circle, the saliency score in that region increases, showing that the model focuses more on the masked area. Please refer to \Cref{app:visual-prompting} for more examples. Table \ref{tab:table2} shows that using visual prompting improves the quality of generated outputs and in reducing hallucinations, thanks to generating higher-quality negative images.

\paragraph{Impact of List Sizes.}

We present the results of LPOI with list sizes of 3, 4, and 5, assessing the impact of the list size on the performance. \Cref{tab:table3} shows that larger list sizes result in improved overall performance, particularly for the Idefics2-8B on the Object HalBench dataset. 
We speculate that this improvement comes from the model being able to learn from a wider range of preference samples, i.e., larger list sizes provide more opportunities for the model to capture fine-grained differences between candidates, leading to a better model performance.

\paragraph{Ablating the DPO Loss.}
The listwise preference loss only utilizes positive text (with multiple masked images). Without the text DPO loss, negative text samples in the dataset are not used, meaning that the model would not learn from any textual preference information (i.e., learn from only image preference information). To demonstrate the effect of incorporating the text DPO loss, we conducted ablation experiments by training the Idefics2-8B model on a 5K training dataset with LPOI (a list size of 3) for 3 epochs. We compared three scenarios: (1) LPOI without the text DPO loss, (2) LPOI with neither the text DPO loss nor the anchor loss, and (3) the full LPOI loss as proposed in this paper. The results, presented in Table \ref{tab:ablation}, show that excluding the DPO loss leads to suboptimal performance compared to using the complete LPOI loss.

\begin{table}[t]
\centering
\small
\setlength{\tabcolsep}{3pt} 
\begin{adjustbox}{max width=1.0\columnwidth}
\begin{tabular}{lccccccc}\toprule
 & \multicolumn{2}{c}{Object HalBench} & \multicolumn{2}{c}{MMHalBench} & \multicolumn{2}{c}{AMBER} \\ 
\cmidrule(lr){2-3} \cmidrule(lr){4-5} \cmidrule(lr){6-7}
Method &  CHAIR$_{s}$ $\downarrow$ & CHAIR$_{i}$ $\downarrow$ & Score $\uparrow$ & HalRate $\downarrow$ & CHAIR$_{s}$ $\downarrow$ & HalRate $\downarrow$ \\ \midrule
Idefics2-8B  & 6.3 & 4.2 & 2.62 & 0.43 & 3.4 & 7.6 \\
+LPOI (without DPO loss)  & 7.7 & 4.6 & 2.56 & 0.44 & 3.3 & 7.4 \\
+LPOI (without DPO, anchor loss)  & 6.0 & 4.1 & 2.50 & 0.45 & 3.5 & 7.5 \\
+LPOI & \textbf{5.7} & \textbf{3.6} & \textbf{2.74 }& \textbf{0.40} & \textbf{2.8} & \textbf{6.4} \\
\bottomrule
\end{tabular}
\end{adjustbox}
\caption{Ablation experiments comparing (1) LPOI without text DPO loss, (2) LPOI without both text DPO and anchor loss, and (3) the full LPOI loss, using the Idefics2-8B model trained for 3 epochs on 5K dataset, using list size of 3.}
\label{tab:ablation}
\end{table}

\paragraph{Results on Different Training Sets.}
We quantitatively compare between DPO, mDPO, and LPOI when training on smaller datasets for Idefics2-8B in \Cref{fig:small-data}. We train them on the subsets with sizes of 5K, 7K, and 10K, repeating the process three times for each subset, and report the average and standard deviation of the GPT-score and hallucination ratio on the MMHalBench benchmark. Our experiments demonstrate a consistent advantage of LPOI over the other methods, both in terms of output quality and  hallucination reduction, across preference datasets of varying sizes.

\paragraph{Qualitative Examples}
Figure \ref{fig:qualitativeexample} presents a comparative analysis of outputs from the LLaVA-v1.5-7B  base model and its fine-tuned variants using DPO, mDPO, and LPOI. For instance, in the first example, where the main factor in hallucination is determining which tomato is the ripest, our model accurately selects the leftmost tomato while other models erroneously choose the rightmost one. The baselines' explanations often contradict what is clearly observable in the image. These results highlight the importance of guiding the model to focus on subtle incremental visual changes. By doing so, our LPOI enables the model to ground its responses more reliably in the image, improving the recognition of fine details and reducing the likelihood of hallucinating common yet irrelevant objects.

\section{Conclusion}
In this work, we addressed the challenge of aligning VLMs with human preferences by proposing LPOI, a novel framework that combines hard negative sampling with listwise ranking. By generating object-aware hard negatives through masking key objects in images and interpolating between them and positive samples, we provide an efficient method for creating listwise preference data without additional annotation cost.
Extensive evaluations on Object HalBench, MMHalBench, and AMBER benchmarks demonstrate that LPOI significantly improves performance by mitigating hallucinations and enhancing multimodal alignment.

\section*{Ethics Statement}
We have used open source models, libraries, datasets, and closed source models for their intended use and license, and not use other than research purposes.

\section*{Limitations}
A potential limitation of our approach is that while we focus on listwise sample generation for the vision and language domain, we do not address other modalities, such as the audio domain. Future work could explore further optimization strategies and extend listwise preference learning to additional modalities, including audio, by adapting similar interpolation strategies to reduce hallucinations in those domains. 
Additionally, the prompts
provided are exclusively in English but it can be expanded to include multiple languages in future iterations. 

\section*{Acknowledgements}

We would like to thank the anonymous reviewers and Professor Chenglin Fan for their valuable feedback. This work was financially supported by Institute of Information \& Communications Technology Planning \& Evaluation (IITP) grant funded by the Korea government (MSIT) (No.~RS-2019-II191082, SW StarLab, No.~RS-2022-II220156, Fundamental research on continual meta-learning for quality enhancement of casual videos and their 3D metaverse transformation, and No.~RS-2021-II211343, Artificial Intelligence Graduate School Program (Seoul National University)), the National Research Foundation of Korea (NRF) grant funded by the Korea government (MSIT) (No.~2023R1A2C2005573), and Basic Science Research Program through the National Research Foundation of Korea (NRF) funded by the Ministry of Education (RS-2023-00274280).

\bibliography{custom}
\onecolumn
\appendix

\section{Experimental Details}
\label{sec:detail}
\textbf{Training setup and hyperparameters} We report the hyperparameters for training LPOI in \Cref{tab:training-detail}.
We fine-tune base models with LoRA adapter with the configuration in \Cref{tab:training-detail}. 

\begin{table}[ht]
    \centering
    \begin{adjustbox}{max width=1.0\columnwidth}
    \begin{tabular}{lccccc}
    \toprule
    Model & LLaVA-v1.5-7B & LLaVA-v1.5-13B & Idefics2-8B \\
    \midrule
    Training epochs & 1 & 1 & 3 \\
    Training set size & 10K & 10K & 10K \\
    Batch size & 64 & 64 & 64   \\
    Optimizer & AdamW & AdamW & AdamW  \\
    Learning rate & 1e-6 & 1e-6 & 5e-6 \\
    Learning rate scheduling & Linear & Linear & Linear  \\
    Mixed precision & FP16 & FP16 & BF16 \\
    LoRA rank & 8 & 8 & 8\\
    LoRA alpha & 8 & 8 & 8\\
    LoRA dropout & 0.0 & 0.0 & 0.0 \\
    \bottomrule

    \end{tabular}
    \end{adjustbox}
\caption{Training hyperparameters for fine-tuning LLaVA-v1.5-7B, LLaVA-v1.5-13B, and Idefics2-8B models.}
    \label{tab:training-detail}
\end{table}

\section{Computational Overhead and Performance Analysis}
We present the training time of DPO, mDPO, and LPOI with the list sizes of 3, 4, and 5 on 5K examples for 1 epoch, measured on an RTX A6000 GPU in Table \ref{tab:trainingtime}. Additionally, we include the number of epochs and scores on the MMHalBench benchmark when trained with the same GPU budget (20 GPU hours), also in Table \ref{tab:trainingtime}. As the list size increases, LPOI introduces computational overhead, but it provides richer signals that help reduce hallucinations, leading to a lower hallucination ratio (See \Cref{tab:table3} ). Moreover, with sufficient optimization time, LPOI outperforms both mDPO and DPO within the same GPU training budget, benefiting from these richer signals.

\begin{table}[h]
\centering
\begin{adjustbox}{max width=\textwidth}
\begin{tabular}{lcccc}
\toprule
Methods & 
\makecell{Time per\\epoch} & 
\makecell{Epochs under\\20 GPU hours} & 
\makecell{MMHalBench\\GPT-Score (↑)} & 
\makecell{MMHalBench\\HalRate (↓)} \\
\midrule
DPO & 2.2 hrs & 9 epochs & 2.29 & 0.51 \\
mDPO & 4.0 hrs & 5 epochs & 2.71 & 0.42 \\
LPOI (list size 3) & 4.5 hrs & 4.5 epochs & -- & -- \\
LPOI (list size 4) & 5.3 hrs & 3.8 epochs & -- & -- \\
LPOI (list size 5) & 6.2 hrs & 3 epochs & \textbf{2.81} & \textbf{0.38} \\
\bottomrule
\end{tabular}
\end{adjustbox}
\caption{Training time per epoch on 5K examples for DPO, mDPO, and LPOI (list sizes 3, 4, 5), using an RTX A6000 GPU, along with the number of epochs and MMHalBench results under the same GPU budget.}
\label{tab:trainingtime}
\end{table}

\section{Extended Benchmark Comparison}
We further evaluated our method (LPOI) and the baselines (DPO and mDPO) on the HallusionBench benchmark \cite{guan2024hallusionbenchadvanceddiagnosticsuite} using Idefics2-8B model, and presented the results in Table \ref{tab:hallusionbench}. LPOI consistently outperforms or matches the baseline methods across most of the metrics.

\begin{table}[h]
\centering
\begin{adjustbox}{max width=\textwidth}
\begin{tabular}{lccccc}
\toprule
\textbf{Model} & \textbf{Question Pair Acc} & \textbf{Figure Acc} & \textbf{Easy Acc} & \textbf{Hard Acc} & \textbf{All Acc} \\
\midrule
Idefics2-8B \citep{laurençon2024mattersbuildingvisionlanguagemodels} & 8.35 & 14.16 & 32.53 & 30.93 & 35.08 \\
+DPO \citep{rafailov2024directpreferenceoptimizationlanguage} & 15.82 & 22.54 & 49.45 & 33.72 & 46.68 \\
+mDPO \citep{wang2024mdpoconditionalpreferenceoptimization} & 16.48 & \textbf{24.28} & 50.33 & 36.05 & 48.45 \\
+LPOI (Ours) & \textbf{17.80} & 23.70 & \textbf{51.65} & \textbf{36.98} & \textbf{49.78} \\
\bottomrule
\end{tabular}
\end{adjustbox}
\caption{Performance comparison between various preference learning methods on HallusionBench benchmark.}
\label{tab:hallusionbench}
\end{table}

\section{Additional Results with Increased Training Data}

We chose to use a 10K subset of Silkie and LLaVA-Instruct-150K for preference fine-tuning by following the experiment setup in mDPO. Furthermore we conducted additional experiments by fine-tuning the Idefics2-8B model on 15K data for 1 epoch, using our method (LPOI) and baselines (DPO, mDPO). The results, presented in Table \ref{tab:larger_dataset_results}, demonstrate that our method consistently outperforms the baselines across most metrics.

\begin{table*}[t]
\centering
\small
\setlength{\tabcolsep}{3pt} 
\begin{adjustbox}{max width=2.5\columnwidth}
\begin{tabular}{lccccccccc}\toprule
 & \multicolumn{2}{c}{Object HalBench} & \multicolumn{2}{c}{MMHalBench} & \multicolumn{4}{c}{AMBER} \\ 
\cmidrule(lr){2-3} \cmidrule(lr){4-5} \cmidrule(lr){6-9}
Method & CHAIR$_{s}$ $\downarrow$ & CHAIR$_{i}$ $\downarrow$ & Score $\uparrow$ & HalRate $\downarrow$ & CHAIR$_{s}$ $\downarrow$& Cover. $\uparrow$ & HalRate $\downarrow$ & Cog. $\downarrow$ \\ \midrule
Idefics2-8B \citep{laurençon2024mattersbuildingvisionlanguagemodels} & 6.3 & 4.2 & 2.62 & 0.43 & 3.4 & 36.5 & 7.6 & 0.4 \\
+ DPO \citep{rafailov2024directpreferenceoptimizationlanguage} & 6.3 & 4.4 & 2.57 & 0.44 & 3.3 & 36.4 & 7.3 & \textbf{0.3} \\ 
+ mDPO \citep{wang2024mdpoconditionalpreferenceoptimization} & 7.7 & 5.0 & 2.74 & 0.41 & \textbf{3.0} & \textbf{37.6} & \textbf{6.8} & \textbf{0.3} \\
+ LPOI (Ours) & \textbf{5.0} & \textbf{3.7} & \textbf{2.75} & \textbf{0.38} & \textbf{3.0} & 36.8 & \textbf{6.8} & \textbf{0.3}\\
\bottomrule
\end{tabular}
\end{adjustbox}
\caption{Performance comparison between various preference learning methods with larger dataset (15K).}
\label{tab:larger_dataset_results}
\end{table*}

\section{Analysis and Ablation of the Verification Module}
For the full 10K dataset with a list size of 5, object detection takes 1,298 seconds (21 minutes), and the verification module takes 18,938 seconds (5.26 hours), averaging 0.166 seconds and 2.43 seconds per data point, respectively. While we reported the version with verification to achieve the best performance, we note that our method performs well even without the verification step, outperforming all baseline methods in this case. To further illustrate this, we conducted an additional experiment using only the object detection module, focusing on a single salient object per image and excluding the verification step, and presented the results in Table \ref{tab:verificationmodule}. Despite this simplification, the LPOI still enables the model to outperform baseline methods like DPO and mDPO across most metrics—especially on hallucination scores, as shown in Table \ref{tab:verificationmodule}. This demonstrates that our approach can maintain strong performance while significantly reducing preprocessing time.

\begin{table*}[t]
\centering
\small
\setlength{\tabcolsep}{3pt} 
\begin{adjustbox}{max width=2.5\columnwidth}
\begin{tabular}{lccccccccc}\toprule
 & \multicolumn{2}{c}{Object HalBench} & \multicolumn{2}{c}{MMHalBench} & \multicolumn{4}{c}{AMBER} \\ 
\cmidrule(lr){2-3} \cmidrule(lr){4-5} \cmidrule(lr){6-9}
Method & CHAIR$_{s}$ $\downarrow$ & CHAIR$_{i}$ $\downarrow$ & Score $\uparrow$ & HalRate $\downarrow$ & CHAIR$_{s}$ $\downarrow$& Cover. $\uparrow$ & HalRate $\downarrow$ & Cog. $\downarrow$ \\ \midrule
Idefics2-8B \citep{laurençon2024mattersbuildingvisionlanguagemodels} & 6.3 & 4.2 & 2.62 & 0.43 & 3.4 & 36.5 & 7.6  & 0.4 \\
+ DPO \citep{rafailov2024directpreferenceoptimizationlanguage} & \textbf{6.0} & 4.2 & 2.48 & 0.45 & 3.5 & 37.4 & 8.1 & \textbf{0.2} \\ 
+ mDPO \citep{wang2024mdpoconditionalpreferenceoptimization} & 7.3 & 5.4 & 2.80 & 0.40 & \textbf{2.7} & 37.7 & 6.2 & \textbf{0.2} \\
+ LPOI (without verification) & \textbf{6.0} & \textbf{4.1} & \textbf{2.86} & \textbf{0.35} & \textbf{2.7} & 36.1 & \textbf{5.9} & \textbf{0.2} \\
\midrule
\midrule
+ LPOI (with verification)	& 5.3 & 3.6 & 2.88 & 0.36 & 2.6 & 36.4 & 5.7 & 0.2 \\
\bottomrule
\end{tabular}
\end{adjustbox}
\caption{Performance of DPO, mDPO, and LPOI on the Idefics2-8B model trained for 3 epochs. LPOI preserves its superiority over the baselines even without verification module. LPOI with verification is included for reference. }
\label{tab:verificationmodule}
\end{table*}

\section{Details on Object Detection Model}
For the object detection component in \Cref{hardnegative generation}, we utilize the Grounding-DINO-Tiny model. Since generating accurate hard negative samples is vital for our pipeline, and precise object detection plays a key role in this process, we evaluate various object detection models to find the most suitable one for our task. Specifically, we compare different versions of Grounding-DINO \citep{liu2024groundingdinomarryingdino}, OwlV2 \citep{minderer2024scalingopenvocabularyobjectdetection}, and YOLO-World \citep{cheng2024yoloworldrealtimeopenvocabularyobject} on a 1k subset of our dataset. The chosen model, with 172 million parameters, effectively detects around 80\% of the key noun objects present in the image.

\section{Details on Visual Prompting}
\label{app:visual-prompting}
\begin{figure*}[ht]
    \centering
    \includegraphics[width=1.0\textwidth]{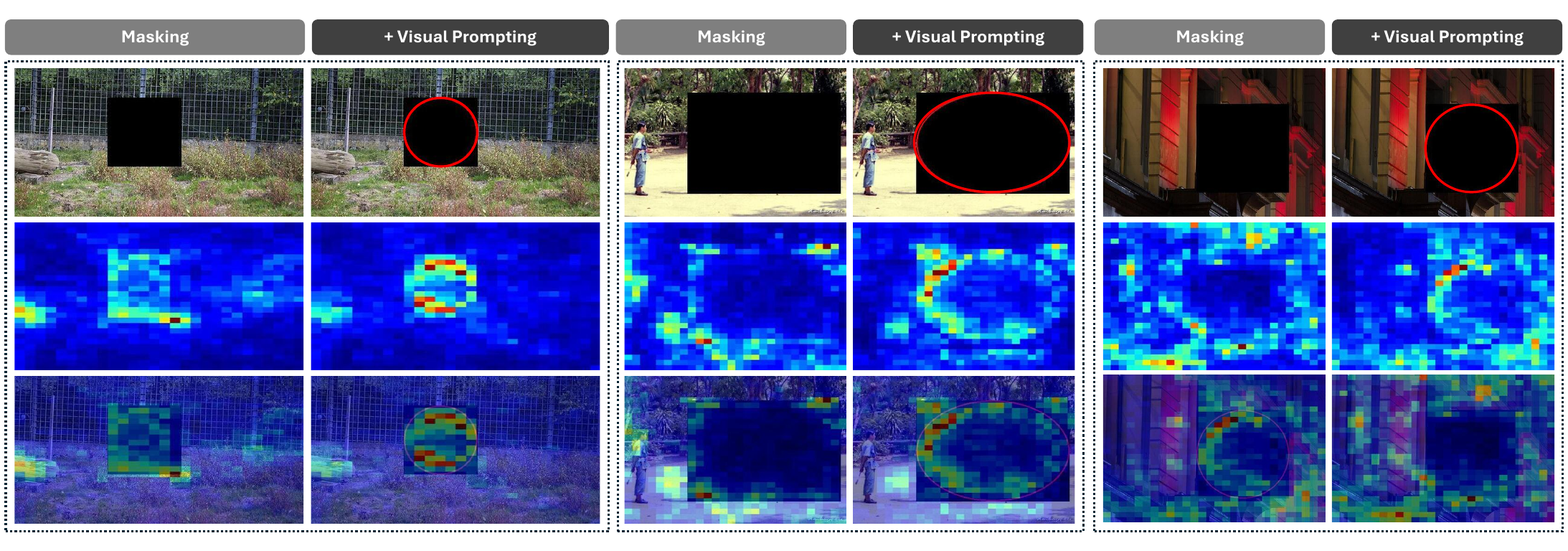}
    \caption{Comparison of saliency maps with or without visual prompting (highlighted in red circle).}
    \label{fig:saliencyappendix}
\end{figure*}

\noindent Figure \ref{fig:saliencyappendix} illustrates 3 more examples of the impact of incorporating an additional visual prompting represented by a red circle in the image, to guide the model's attention toward the region of interest.  
In each group, the left column displays an image from our dataset with only the applied mask, its corresponding saliency map, and an overlap visualization of the two. The right column shows the same image, but with the visual prompt added by circling the masked area.

\section{Qualitative Analysis}

\begin{figure*}[b]
    \centering
    \includegraphics[width=1.0\textwidth]{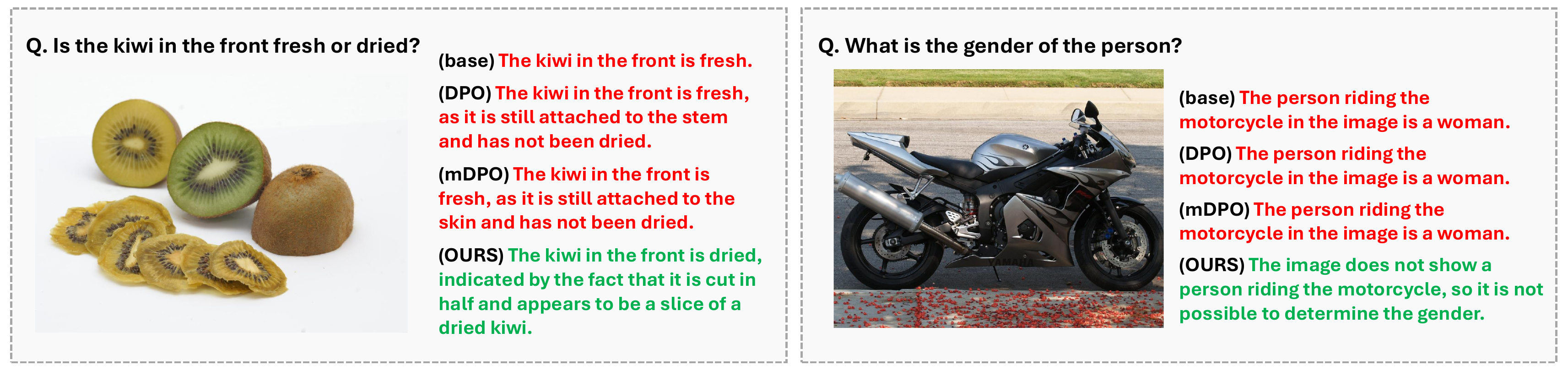}
    \caption{Qualitative results of the base model (LLaVA-v1.5-7B) its variants optimized with DPO, mDPO, and LPOI(Ours). \textbf{\textcolor{darkgreen}{Correct answers}} and \textbf{\textcolor{red}{hallucinations}} are highlighted.}
    \label{fig:qualitativeresult2}
\end{figure*}

\noindent We provide additional examples generated by the fine-tuned models using DPO, mDPO, and LPOI (ours). 
In the first example, shown at the left, all models except ours mistakenly claim that the kiwi in the foreground, which is dried into chips, is fresh. In the third example at the right, the image shows a motorcycle without a rider. When asked to determine the gender of the person riding the motorcycle, our model correctly states that no person is visible, while the other models erroneously identify a woman as the rider. These examples highlight how our method reduces common hallucinations in vision-language models, such as the false assumption of co-occurring objects, the failure to recognize subtle object features or the provision of answers to questions that cannot be derived from the image alone.

\section{Details on Human Evaluation}
\label{sec:human}
\begin{figure*}[t]
    \centering
    \includegraphics[width=0.8\textwidth]{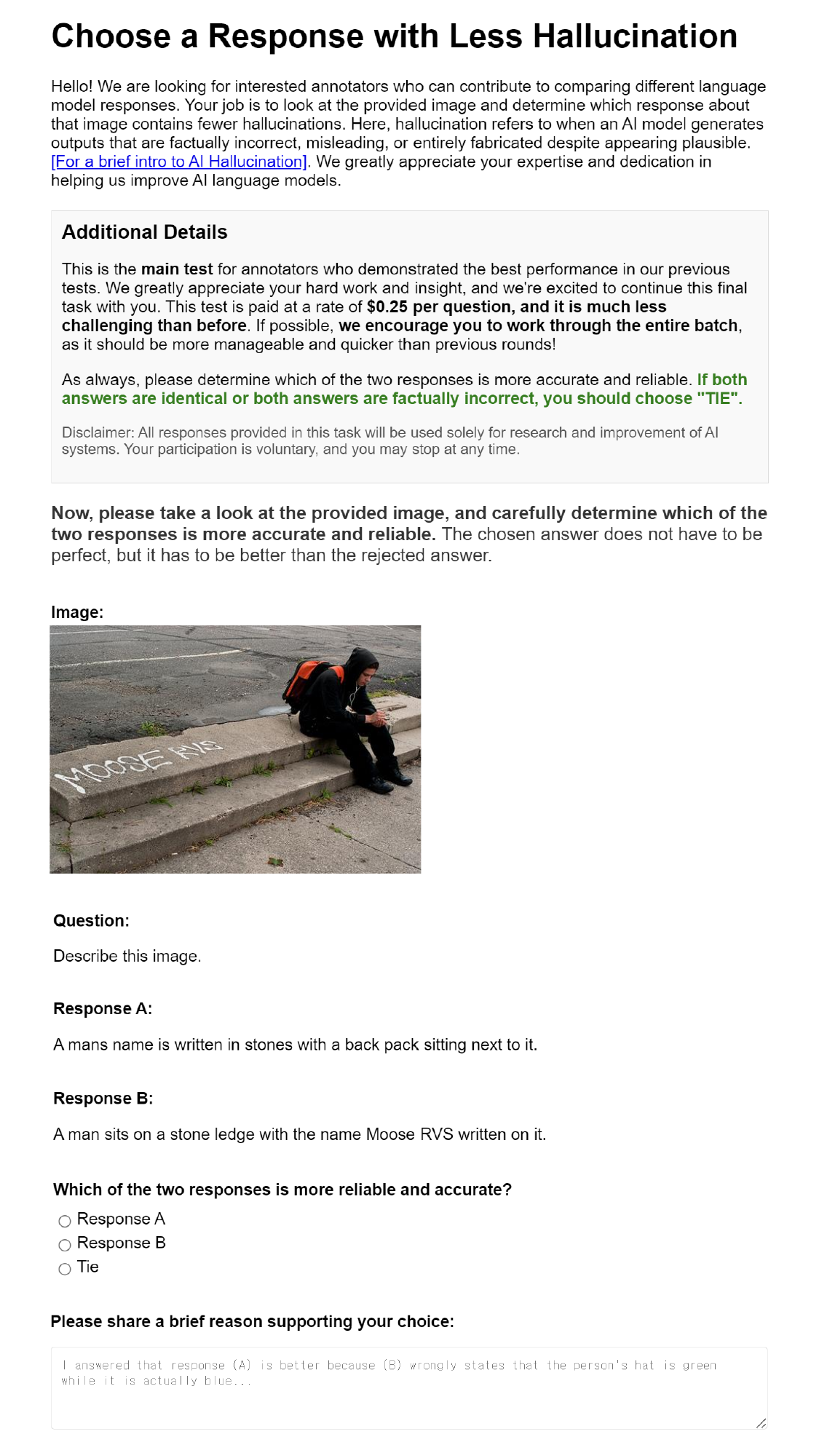}
    \caption{User interface and instruction for human evaluation.}
    \label{fig:mturk}
\end{figure*}

\Cref{fig:mturk} shows the user interface where annotators select the less hallucinatory response between two answers generated by mDPO and LPOI (ours). Each worker is presented with two responses generated by the Idefics2-8B model: one fine-tuned using mDPO or DPO, and the other using the LPOI method. Workers are instructed to select the response they consider more accurate and reliable based on the visual information in the image. If the responses are identical or both factually incorrect, workers are asked to choose the 'tie' option. The workers' answers are then aggregated using a majority vote. To prevent bias, the order of the responses (Response A and Response B) is shuffled for each datapoint, and workers must also provide justifications for their selections. These justifications are reviewed to ensure the reliability and consistency of the answers, which are then used to validate the integrity of the evaluation process.

\end{document}